\documentclass[10pt,twocolumn,letterpaper]{article}

\usepackage{wacv}                         

\usepackage{multirow}
\usepackage{array}
\usepackage{adjustbox}
\newcolumntype{R}{>{\raggedleft\arraybackslash}p{2.8em}}

\usepackage{stfloats}
\definecolor{wacvblue}{rgb}{0.21,0.49,0.74}
\usepackage[pagebackref,breaklinks,colorlinks,allcolors=wacvblue]{hyperref}

\def\wacvPaperID{724}
\def\confName{WACV}
\def\confYear{2027}

\title{Shifting Focus by Reliability: Geometric Fusion Modulation for Robust RGB-T Detection}

\author{Liuyin Shi\\
McMaster University\\
{\tt\small shil9@mcmaster.ca}
\and
Chengqi Li\\
McMaster University\\
{\tt\small lic222@mcmaster.ca}
\and
Wenbo He\\
McMaster University\\
{\tt\small hew11@mcmaster.ca}
}

\begin{document}
\maketitle

\begin{abstract}
RGB--thermal (RGB--T) object detectors combine visible and thermal imagery for robust detection, but signal corruption in either stream can substantially degrade accuracy. Restoration-based methods incur significant latency and do not consistently improve downstream accuracy, raising a more direct question: rather than restoring corrupted inputs, can we efficiently adapt the detector to their varying
signal quality? Conventional PEFT offers an efficient route, but its input-independent corrections cannot track diverse test-time corruptions. Our controlled corruption sweeps show that corruption types have similar effects near clean conditions but diverge increasingly with severity, motivating corruption-aware, per-input adaptation. We therefore propose \textbf{PG-FM} (Prompt-Guided Feature Modulation), which adapts RGB--T object detectors to the estimated corruption type and severity of each modality. Under controlled RGB--T corruption simulations, PG-FM preserves clean-set
performance, improves mAP$_{50}$ by up to $20.8$ points on FLIR and $19.9$
points on KAIST, and transfers without retraining to unmodified, naturally acquired
external RGB--T datasets.
\end{abstract}

\section{Introduction}
\label{sec:intro}
\begin{figure}[t]
  \centering
  \includegraphics[width=\linewidth,height=7.1cm,keepaspectratio]{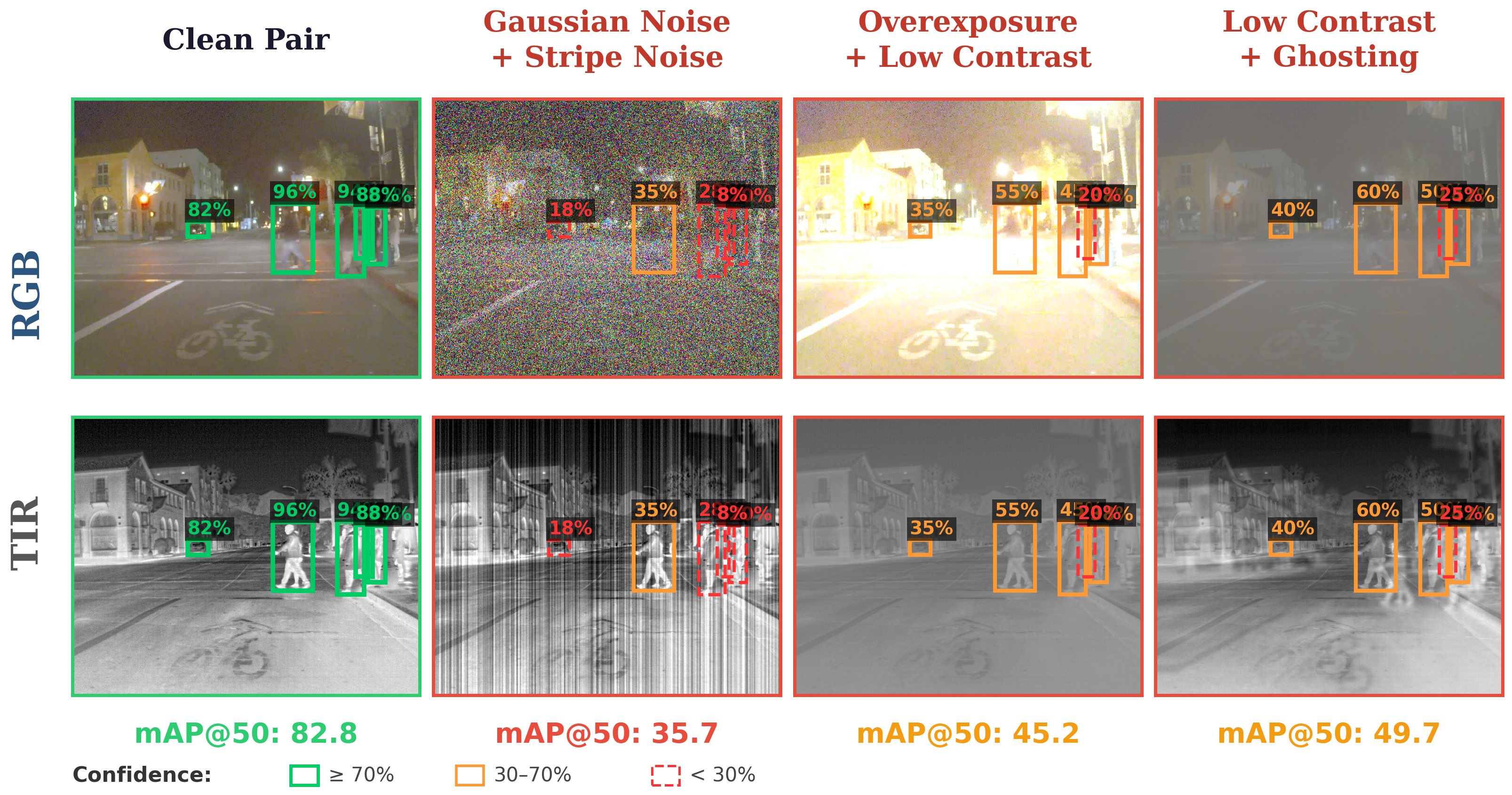}
  \caption{A clean-trained RGB-T detector degrades under simulated sensor corruptions, with the accuracy loss depending on corruption type---motivating per-sample fusion adaptation.}
  \label{fig:corruption_impact}
\end{figure}

RGB-thermal (RGB--T) fusion underpins robust object detection in domains such as autonomous driving and pedestrian detection, combining the visible and thermal-infrared (TIR) modalities to sustain detection where either modality alone is unreliable. In practice, state-of-the-art RGB-T detection models optimized on healthy signal input, degrade markedly in the presence of signal corruptions: thermal sensors, particularly uncooled sensor cores, exhibit elevated noise and reduced contrast on reflective surfaces ~\cite{amon2009evaluation,li2015intrusion}, while the visible channel is affected by motion blur, precipitation, glare, and low-light noise, and both modalities may fail outright under hardware faults (\cref{fig:corruption_impact})~\cite{10817087}. Such signal corruption degrades the affected modality’s features, which then propagate through fusion to impair detection accuracy.

A natural response to signal corruption is to restore the corrupted modality before downstream fusion detection. Generative restoration, however, introduces substantial computational overhead and may produce hallucinated structures that mislead the downstream detector~\cite{kim2025exploiting,NEURIPS2022_1cac8326}. We instead adapt the fusion process directly: rather than reconstructing the corrupted signal, the detector should reduce its contribution and shift focus toward the more reliable modality while retaining any complementary information that remains. Parameter-efficient fine-tuning (PEFT) provides an efficient means of adapting pre-trained models ~\cite{hu2022lora,gao2024clip,perez2018film}, but conventional PEFT learns a single input-independent correction fitted to the training corruption distribution and therefore cannot respond to sample-specific reliability changes~\cite{taori2020measuring,li2024flat,mai2025lessons}. Our controlled sweeps make this limitation explicit: corruption effects are
similarly mild near clean inputs but increasingly diverge by type as
severity rises. Restoration does not consistently improve downstream
detection and adds substantial latency, while static PEFT remains tied to
the training corruption taxonomy and transfers poorly to unseen types.

Motivated by this observation, we propose \textbf{PG-FM (Prompt-Guided Feature Modulation)}, which conditions fusion-detection adaptation on modality-specific geometric prompts encoding both corruption type and severity. The Prompt Generator maps spatial and signal-energy features to a Poincar\'e ball, encoding per-modality signal health status, with type by angular direction and branch-relative severity by geodesic depth while enforcing cross-scene consistency. A Reliability Mining Module then converts the RGB and TIR prompts into per-sample PEFT corrections for the fusion-detection layers. This design adapts fusion towards the more reliable modality without explicitly restoring either input, retraining the detector backbone, or discarding the corrupted modality. Our main contributions are:

\begin{itemize}
    \item \textbf{Type-Severity Corruption Geometry.}
    We propose a cross-scene-consistent prompt geometry that organizes
    corruption type by angular direction and orders within-type severity by
    prototype-relative geodesic depth, preserving branch separation under
    high-severity dispersion.
    \item \textbf{Prompt-Conditioned Fusion Adaptation.}
    We translate the RGB and TIR prompt fields into dynamic corrections for
    downstream fusion detector layers, while reliability-weighted identity regularization
    preserves clean behaviour and releases adaptation as prompt-inferred
    reliability decreases.
    \item \textbf{Controlled Robustness and External Transfer.}
      We introduce a RGB--T corruption simulation and evaluation protocol spanning diverse corruption types and test scenarios, motivated by sensor physics of both modalities. We evaluate robustness spanning seen, unseen, mixed, and cross-operator tests, together with zero-shot evaluation on naturally acquired external datasets.
\end{itemize}

\section{Related Work}

\noindent\textbf{Robust RGB--T detection.}
 Robust RGB-T detection maintains reliable fusion when one or both modalities are corrupted. Restoration-assisted methods recover degraded inputs before detection ~\cite{NEURIPS2022_1cac8326,kim2025exploiting}, but add substantial computational cost and may introduce task-harmful artifacts ~\cite{kim2025hallugen,NEURIPS2024_f29369d1}. Other approaches reweight modalities~\cite{yuan2024c}, model shared and modality-specific representations under missing inputs ~\cite{he2023multispectral,ma2022multimodal}, or mitigate training-time modality imbalance~\cite{10817087,TIAN20251}. PG-FM instead keeps the detector frozen and adapts fusion at inference from an explicit per-input reliability representation.
 
\noindent\textbf{Input-conditioned PEFT.}
PEFT methods (\eg LoRA, FiLM, adapters)~\cite{hu2022lora,perez2018film,gao2024clip,pfeiffer2020adapterhub} adapt frozen backbones and extend to multimodal alignment~\cite{11133673}. For RGB-T fusion without input-conditioned control, however, PEFT learns a single static correction biased toward the training corruption distribution, so robustness becomes distribution-dependent even when clean accuracy is preserved~\cite{taori2020measuring,li2024flat}. Dynamic adapter selection has been explored~\cite{Zhou_2024_CVPR}, but prior methods lack an interpretable corruption representation for per-input modulation in RGB-T fusion.

\noindent\textbf{Reliability and geometric conditioning.}
No-Reference Image Quality Assessment (NR-IQA) models extract degradation-aware features, but their scalar
perceptual scores can vary with scene content ~\cite{9156687,10655501,shi2024transformer,10030129}. PG-FM instead learns a cross-scene control geometry that separates corruption type from branch-relative severity. Hyperbolic embeddings provide increasing metric capacity with depth ~\cite{nickel2017poincare,ganea2018hyperbolic, khrulkov2020hyperbolic}, matching the observed expansion of type-dependent corruption effects at high severity.

\section{Method}
\subsection{Problem Formulation}
\label{sec:problem}

Let $x=(x^{\mathrm{rgb}},x^{\mathrm{tir}})$ be a spatially aligned RGB--T pair and $f_\theta$ a detector comprising modality-specific feature extractors, cross-modal fusion layers, and a detection head predicting
targets $y$. At test time, one or both modalities may be corrupted as $\tilde{x}=\mathcal{T}(x;\delta,s)$, where $\delta$ denotes corruption type and $s\in[0,1]$ its nominal severity, with $s=0$ denoting the clean
pair. We study signal corruption under preserved inter-modal registration; geometric misalignment is evaluated separately in \cref{sec:generalization}.

PG-FM keeps $f_\theta$ frozen and does not reconstruct either input. Instead, it learns an input-conditioned correction
$\Delta\theta_\phi(\tilde{x})$ applied only to the fusion layers:
\begin{equation}
  \min_{\phi}\;
  \mathbb{E}_{(\tilde{x},y)}
  \left[
    \mathcal{L}_{\mathrm{det}}\!\left(
      f_{\theta+\Delta\theta_\phi(\tilde{x})}(\tilde{x}),y
    \right)
  \right].
  \label{eq:problem}
\end{equation}
Reliable inputs should remain close to the frozen detector, whereas stronger
within-branch degradation should release greater adaptation capacity. We
implement this stability--plasticity behaviour through a prompt-conditioned
identity prior whose weight decreases monotonically with inferred
degradation. This constrains adaptation capacity rather than imposing a hard
ordering on the realized correction norm, which also depends on corruption
type, spatial evidence, and detector response.

\subsection{Overview of PG-FM}

\begin{figure}[t]
  \centering
  \includegraphics[width=\linewidth]{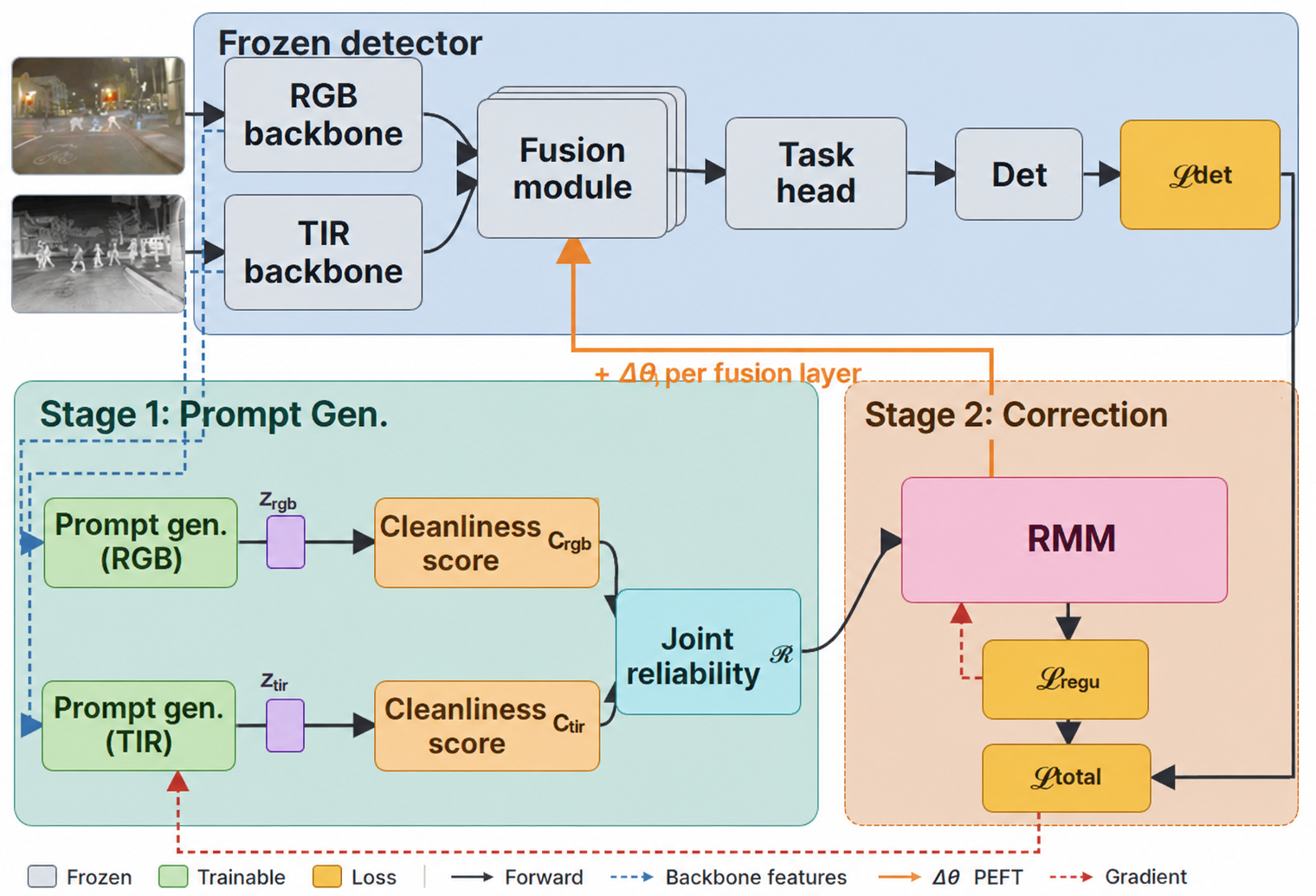}
    \caption{\textbf{PG-FM overview.}
    Stage~1 infers modality-specific type--severity prompt fields from RGB/TIR
    features. Stage~2 converts these prompts into identity-centered,
    input-conditioned corrections for the fusion-detection layers.}
  \label{fig:pgfm_overview}
\end{figure}

\begin{figure}[t]
  \centering
  \includegraphics[width=\linewidth]{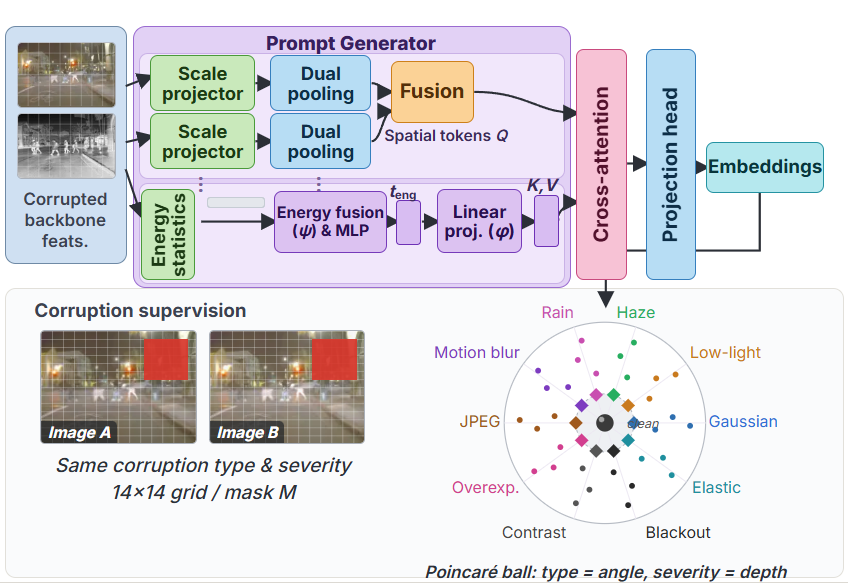}
    \caption{\textbf{Prompt Generator.}
    Multi-scale spatial features and absolute-energy statistics are fused into
    per-patch Poincar\'e embeddings. Angular direction selects a corruption
    branch, while prototype-relative depth orders severity within that branch.}
  \label{fig:pg_architecture}
\end{figure}

PG-FM realises the objective above in two stages
(\cref{fig:pgfm_overview}). Stage~1 learns one Prompt Generator per
modality, mapping both RGB/TIR encoders' multi-scale spatial and signal-energy features to
per-patch numerical prompts on the Poincar\'e ball
(\cref{fig:pg_architecture}). Angular direction represents corruption type,
while prototype-relative geodesic depth orders severity within each type
branch. Corruption type, severity, and spatial masks supervise Stage~1 only;
inference requires only the RGB--T input pair.

Stage~2 freezes the detector and both Prompt Generators. A Reliability
Mining Module (RMM) then converts the RGB and TIR prompt fields into
layer-wise FiLM or LoRA corrections for the fusion-detection layers. The full
prompt field determines the content of each correction, while a pooled
trunk-relative statistic controls how strongly the generated parameters are
regularized toward identity.

\subsection{Stage~1: Corruption Prompt Learning}
\label{sec:stage1}

Stage~1 trains one modality-specific Prompt Generator $\mathrm{PG}_m$, $m\in\{\mathrm{rgb},\mathrm{tir}\}$, for estimating per-modality signal quality.

\textbf{Signal-Energy Token.}\quad
Each backbone feature map $\mathcal F_s$ is projected to width $d$ and
average/max pooled to a $14{\times}14$ grid, yielding $N{=}196$ spatial
tokens. Here, $\mu_s,\sigma_s\in\mathbb R^d$ are channel-wise spatial
moments, while the GAP/GMP descriptors are projected to the energy-token
width.
\begin{equation}
\begin{aligned}
\mathbf e_s
  &= [\,\mu_s;\sigma_s;
      \hat\psi_{\rm a}(\operatorname{GAP}(\mathcal F_s));
      \hat\psi_{\rm m}(\operatorname{GMP}(\mathcal F_s))\,],\\
t_{\rm eng}
  &= \psi\!\left(
      [\,\mathbf e_1;\ldots;\mathbf e_S;
      \log(1+\|[\mathbf e_1;\ldots;\mathbf e_S]\|)\,]\right),\\
\mathcal F'
  &= \mathcal F+\mathbf P+
     \mathbf 1_N\eta(t_{\rm eng})^\top .
\end{aligned}
\label{eq:energy_fusion}
\end{equation}
Here, $\mathbf P$ is the positional embedding and $\eta$ is the learned
value/output projection. The last term is implemented by attention to one
energy token; since its single-key softmax is always one, it acts as global
residual conditioning rather than spatial selection. Self-attention, an FFN,
and a two-layer head then produce
$\mathbf v_n\in\mathbb R^{d'}$ for each spatial token.

\textbf{Type--Severity Geometry on the Poincar\'e Ball.}\quad
Each $\mathbf v_n$ is mapped to the curvature-one Poincar\'e ball by
\begin{equation}
\begin{aligned}
\mathbf z_n
  &= \exp_{\mathbf0}(\mathbf v_n)
   = \tanh\!\left(\frac{\|\mathbf v_n\|}{2}\right)
     \frac{\mathbf v_n}{\|\mathbf v_n\|},\\
d_{\mathbb D}(\mathbf0,\mathbf z_n)
  &= 2\operatorname{atanh}\|\mathbf z_n\|
   = \|\mathbf v_n\|,
\end{aligned}
\label{eq:expmap}
\end{equation}
where $\exp_{\mathbf0}(\mathbf0)=\mathbf0$. Thus, prompt direction routes corruption type, while prototype-relative
depth orders within-branch severity. This separation preserves scene-dependent variation instead of
forcing samples with the same corruption to share one coordinate
~\cite{ganea2018hyperbolic,nickel2017poincare,khrulkov2020hyperbolic,ermolov2022hyperbolic}.

For modality $m$, the geometry comprises a clean trunk $\|\mathbf z\|\leq r^*$ and $K_m$ angular branches anchored by fixed prototypes $\mathbf p_{m,k}=\exp_{\mathbf0}(\rho^*\mathbf e_{m,k})$, where $\rho^*=2\operatorname{atanh}(r^*)$. Fixed prototype directions separate angular type routing from learned radial expansion. The unit directions $\{\mathbf e_{m,k}\}_{k=1}^{K_m}$ are obtained once by maximizing their minimum pairwise angular separation and are held fixed throughout training.

For each nonzero prompt, we define the nearest angular branch
\[
\hat{k}
=
\arg\max_k
\cos(\mathbf z_m,\mathbf p_{m,k}).
\]
A prompt is decoded as clean when $\|\mathbf z\|\leq r^*$ and otherwise as
corruption type $\hat{k}$. Its prototype-relative branch depth is
\begin{equation}
  g(\mathbf{z})
  \triangleq d_{\mathbb{D}}(\mathbf{p}_{\hat{k}},\,\mathbf{z})
  = \operatorname{arcosh}\!\left(
      1{+}\frac{2\,\|\mathbf{p}_{\hat{k}}{-}\mathbf{z}\|^2}
             {(1{-}\|\mathbf{p}_{\hat{k}}\|^2)\,
              (1{-}\|\mathbf{z}\|^2)}
    \right).
  \label{eq:geodesic}
\end{equation}
At $\mathbf z=\mathbf0$, the prompt is clean and no corruption type is
decoded. Training enforces branch agreement and depth consistency for scene-different, corruption-identical pairs while allowing residual content-dependent variation (\cref{fig:stage1_montage}). Because prototype-relative distance can also reflect residual angular
routing error, we interpret $g(\mathbf z)$ as a learned within-branch
severity coordinate rather than a strictly angle-independent scalar.
It is distinct from the pooled trunk-relative radius used in Stage~2 to control the identity prior.

\textbf{Training Procedure.}\quad
Stage~1 follows the controlled corruption protocol in \cref{sec:protocol}. For each image, we sample a corruption type $\delta_k$, severity $s\sim\mathcal U(0,1)$, and binary patch mask $\mathcal M\in\{0,1\}^{14\times14}$. The mask provides spatially local supervision in Stage~1, whereas Stage~2 uses full-image corruptions. Each mini-batch of $B$ images forms $B/2$ scene-different pairs sharing
$(\delta_k,s,\mathcal M)$, providing the supervision for $\mathcal L_{\rm pair}$.

We denote the set of clean patches as
$\mathcal{C}{=}\{n:\mathcal{M}^n{=}0\}$ (unmasked regions
receiving no corruption) and corrupted patches as
$\mathcal{D}{=}\{n:\mathcal{M}^n{=}1\}$ (masked regions
where $\mathcal{T}$ is applied). The total loss
\begin{equation}
  \mathcal{L}_{\mathrm{S1}}
    = \lambda_g\,\mathcal{L}_{\mathrm{grav}}
    + \lambda_b\,\mathcal{L}_{\mathrm{push}}
    + \lambda_t\,\mathcal{L}_{\mathrm{type}}
    + \lambda_r\,\mathcal{L}_{\mathrm{rank}}
    + \lambda_p\,\mathcal{L}_{\mathrm{pair}}
  \label{eq:stage1_loss}
\end{equation}
enforces the trunk-and-branch topology through five terms.
\textit{(i) Boundary.}\;
Clean prompts remain inside the trunk, while corrupted prompts are pushed
beyond the prototype radius:
\begin{align}
  \mathcal{L}_{\mathrm{grav}}
  &= \frac{1}{|\mathcal{C}|}
     \sum_{n\in\mathcal{C}}
     \max\!\bigl(0,\;
       \|\mathbf{z}^n\| - r^*
     \bigr), \label{eq:grav}\\
  \mathcal{L}_{\mathrm{push}}
  &= \frac{1}{|\mathcal{D}|}
     \sum_{n\in\mathcal{D}}
     \max\!\bigl(0,\;
       \rho^* - d_{\mathbb{D}}(\mathbf{0},\mathbf{z}^n)
     \bigr). \label{eq:push}
\end{align}

\textit{(ii) Type and Severity.}\;
For $n\in\mathcal D$, a cosine-softmax assigns the prompt direction to its corruption prototype:
\begin{equation}
  \mathcal{L}_{\mathrm{type}}
  = -\frac{1}{|\mathcal{D}|}
    \sum_{n\in\mathcal{D}}
    \log
    \frac{\exp\!\bigl(
      \hat{\mathbf{z}}^{n\top}\hat{\mathbf{p}}_k/\tau
    \bigr)}
    {\sum_{j=1}^{K}\exp\!\bigl(
      \hat{\mathbf{z}}^{n\top}\hat{\mathbf{p}}_j/\tau
    \bigr)},
  \label{eq:type}
\end{equation}
where hats denote $\ell_2$ normalisation. Since $\mathcal L_{\mathrm{type}}$ depends only on direction, it does not directly supervise depth.

Severity is enforced by ranking same-type pairs within a training batch.
For two scene-different pairs $(i,j)$ with the same type $\delta_k$ and
severities $s_i>s_j$, define
\[
\mathcal D_i^+
=
\left\{
(q,n):
q\in\{1,2\},\;
n\in\mathcal D,\;
d_{\mathbb D}(\mathbf0,\mathbf z_{i,q}^n)>\rho^*
\right\},
\]
and
\[
\bar g_i
=
\frac{1}{|\mathcal D_i^+|}
\sum_{(q,n)\in\mathcal D_i^+}
g(\mathbf z_{i,q}^n).
\]
Pairs for which either $\mathcal D_i^+$ or $\mathcal D_j^+$ is empty do
not contribute to $\mathcal L_{\mathrm{rank}}$. The ranking loss
\begin{equation}
  \mathcal{L}_{\mathrm{rank}}
  = \max\!\bigl(0,\;
    \kappa(s_i{-}s_j) - (\bar{g}_i{-}\bar{g}_j)
    \bigr),
  \label{eq:rank}
\end{equation}
encourages depth gaps to scale with severity gaps, with slope $\kappa$.

\textit{(iii) Cross-Scene Consistency.}\;
For each modality, with the modality subscript omitted for clarity, we pair
scene-different samples sharing the same corruption type, severity, and
spatial mask $(\delta_k,s,\mathcal M)$, and minimize the absolute difference
between corresponding corrupted-patch depths:
\begin{equation}
  \mathcal{L}_{\mathrm{pair}}
  =
  \frac{1}{|\mathcal D|}
  \sum_{n\in\mathcal D}
  \left|
    g(\mathbf z_i^n)-g(\mathbf z_j^n)
  \right|.
  \label{eq:pair}
\end{equation}
This depth-only term complements $\mathcal L_{\mathrm{type}}$ and is
isolated by the \emph{w/o $L_{\rm pair}$} controls in
\cref{tab:ablation,tab:external_transfer}.

\subsection{Stage~2: Reliability Mining Module}
\label{sec:stage2}

Stage~2 freezes the detector and both Prompt Generators and trains only the Reliability Mining Module. Full-image corruptions follow protocol~(ii), with clean pairs included to preserve the frozen detector on unmodified inputs.

\textbf{Prompt-Conditioned PEFT Generation.}\quad
For modality $m\in\{\mathrm{rgb},\mathrm{tir}\}$, the frozen Prompt Generator produces per-patch embeddings
$\{\mathbf z_m^n\}_{n=1}^{N}$. We map them to the tangent space at the origin:
\begin{equation}
  \mathbf{t}_m^n
  = \log_{\mathbf{0}}(\mathbf{z}_m^n)
  = \frac{2\operatorname{atanh}(\|\mathbf{z}_m^n\|)}
         {\|\mathbf{z}_m^n\|}\,\mathbf{z}_m^n\,,
  \label{eq:logmap}
\end{equation}
where $\log_{\mathbf0}(\mathbf0)=\mathbf0$ by continuity.
After a learned projection $\pi:\mathbb R^{d'}\!\rightarrow\mathbb R^d$, the RGB and TIR tangent
tokens form $\mathbf Z\in\mathbb R^{2N\times d}$, preserving modality-specific spatial reliability evidence. The RMM maintains one learnable query $\mathbf q_\ell$ per target fusion layer; self-attention
coordinates the layer-wise updates, and cross-attention to $\mathbf Z$ selects relevant spatial and modality evidence. A linear head maps each refined query to an identity-centred PEFT update.
For FiLM,
$h'_\ell=(1+\Delta\gamma_\ell)\odot h_\ell+\Delta\beta_\ell$;
for LoRA,
$W'_\ell=W_\ell+B_\ell A_\ell$.
Thus, $\Delta\boldsymbol{\theta}_\ell=\mathbf0$ exactly recovers the frozen
fusion layer.

\textbf{Reliability-Weighted Identity Regularization.}\quad
The RMM should preserve the frozen fusion pathway on reliable inputs while
retaining sufficient capacity to correct severely degraded features. We
therefore scale an identity-centred $\ell_2$ penalty by the joint reliability
inferred from the modality prompt fields. Near-trunk prompts impose a strong
identity prior, whereas outward prompts relax the penalty and progressively
release adaptation capacity. We summarise the per-patch prompt field for
modality $m$ using depth-weighted tangent pooling:
\begin{equation}
  \bar{\mathbf{v}}_m
  = \sum_{n=1}^{N}
    \frac{\|\mathbf{t}_m^n\|}
         {\sum_{j}\|\mathbf{t}_m^j\|}\;
    \mathbf{t}_m^n,
  \qquad
  \bar{r}_m
  = \tanh\!\bigl(\|\bar{\mathbf{v}}_m\|/2\bigr),
  \label{eq:pooled_radius}
\end{equation}
When the denominator vanishes, we set
$\bar{\mathbf v}_m=\mathbf0$.
Here, $\bar r_m$ is a trunk-relative degradation radius rather than a
globally calibrated cross-type severity score. It controls only the
identity-prior coefficient $c_m$; the RMM still receives the complete
angular, spatial, and depth prompt field. Consequently, equal
$\bar r_m$ implies the same identity-prior strength but not necessarily the
same realised correction.
\begin{equation}
  c_m =
  \sigma\!\bigl({-}\beta\,(\bar r_m/r^*{-}1)\bigr),
  \label{eq:cleanliness}
\end{equation}
with $c_m{\approx}1$ at shallow depth
($\bar{r}_m{\lesssim}r^*$) and $c_m{\to}0$ as tokens push
deeper. The Stage~2 objective combines the detection loss
$\mathcal{L}_{\mathrm{det}}$ with a dynamic-weight
$\ell_2$ penalty on the squared Frobenius norm of the generated PEFT parameters, weighted by the inferred joint reliability:
\begin{equation}
  \mathcal{L}_{\mathrm{S2}}
  =
  \mathcal{L}_{\mathrm{det}}
  +
  \lambda_{\mathrm{id}}
  \sum_{\ell=1}^{L}
  \frac{1}{B}\sum_{b=1}^{B}
  \mathcal R^{(b)}
  \left\|
    \Delta\boldsymbol{\theta}_{\ell}^{(b)}
  \right\|_{\mathrm F}^{2},
  \label{eq:stage2_loss}
\end{equation}
where
$\mathcal R^{(b)}
=c_{\mathrm{rgb}}^{(b)}c_{\mathrm{tir}}^{(b)}+\epsilon_0$
is the joint reliability weight, and $\epsilon_0>0$ prevents the identity
constraint from vanishing completely. When both modalities are reliable,
$\mathcal R\approx1+\epsilon_0$ and the regularizer keeps the generated
correction near identity. When either modality is severely degraded,
$\mathcal R$ approaches $\epsilon_0$, relaxing the identity constraint and
allowing stronger adaptation.

\subsection{RGB-T Corruption Simulation Protocol}
\label{sec:protocol}

We define a controlled protocol over paired RGB--T inputs. TIR operators act in detector input space and approximate documented microbolometer/IRFPA failures~\cite{s23208424,liu2025deal}; RGB operators follow photographic and adverse-weather models ~\cite{hendrycks2019robustness,garg2006photorealistic}. Continuous corruption simulation of types defined in \cref{tab:corruptions} interpolate from the clean input at $s=0$ to a predefined maximum simulated failure at $s=1$; blackout is binary. These are sensor-physics motivated simulations, not calibrated replicas of a specific sensor. 

We evaluate: \textit{(i)} single-modality, fixed-type severity sweeps with
the complementary modality clean; \textit{(ii)} independently and uniformly
sampled RGB/TIR types and severities for the Robust setting; and
\textit{(iii)} unseen-corruption sweeps or mixed corruption tests applied to
one modality with the other clean. All methods share identical corruption
draws and seeds; exact implementation details are supplementary.

\begin{table}[t]
    \centering
    \scriptsize
\caption{\textbf{Simulated sensor corruptions.}
Thermal types approximate microbolometer/IRFPA failures; visible types follow standard models. Each maps to one Poincar\'e prototype; continuous $s$ interpolates from clean to a predefined maximum, while blackout is binary.}
    \label{tab:corruptions}
    \setlength{\tabcolsep}{4pt}
    \begin{tabular}{@{}cp{0.78\columnwidth}@{}}
        \toprule
        \textbf{Mod.} & \textbf{Corruption Types ($K$)} \\
        \midrule
        RGB\,(10 types) & Gaussian noise, low-light/dark, atmospheric haze, rain streaks, motion blur, JPEG compression, overexposure, contrast reduction, total blackout, elastic distortion \\[2pt]
        TIR\,(9 types) & De-focus blur, Gaussian noise, stripe noise streaks, fixed-pattern non-uniformity, contrast reduction, ghosting artifacts, dead pixels, sensor solar burn artifacts, total blackout \\
        \bottomrule
    \end{tabular}
\end{table}

\section{Experiments}
\label{sec:experiments}

\subsection{Setup}
\label{sec:setup}

\textbf{Datasets and metrics.}\quad
We evaluate controlled corruption robustness on FLIR-Aligned
~\cite{zhang2020multispectral} using mAP$_{50}$ and Precision, and on
KAIST~\cite{7298706} using mAP$_{50}$ and LAMR over
FPPI$\in[10^{-2},10^{0}]$ for the All/Day/Night subsets. For external
transfer, models trained only on FLIR-Aligned are evaluated without
target-domain optimisation or added corruption on the unmodified
LYNRED-MDS~\cite{11533396}, M3FD~\cite{liu2022target}, and
ATR-UMOD~\cite{chen2025fusion} datasets.

\textbf{Models and controls.}\quad
We instantiate PG-FM on ICAFusion~\cite{shen2024icafusion}, generating FiLM or LoRA corrections for its Q/K/V/O fusion projections ~\cite{perez2018film,hu2022lora}. Backbone RGB--T detectors are frozen and  provide contextual references, while static PEFT, capacity-matched Zero-Prompt, alternative prompts, generic feature conditioning, and direct reliability gating follow the common protocol in \cref{sec:ablations}. BioIR~\cite{NEURIPS2025_73ba81c7} and PromptIR~\cite{potlapalli2023promptir} provide restore-then-detect controls: both modalities are restored independently before the resulting RGB--T pair is passed to the same detector without target-domain training.

\textbf{Implementation.}\quad
Stage~1/2 train for 25/50 epochs with AdamW; timings use an NVIDIA
RTX~5070~Ti. PG-FM adds $5.0/6.9$\,M trainable parameters for FiLM/LoRA.
On ICAFusion, FiLM/LoRA add $3.2/4.1\%$ GFLOPs and $10.5/12.3\%$ latency;
additional backbone results and costs are supplementary.

\subsection{Controlled Robustness and Zero-Shot Transfer}
\label{sec:main_results}

\begin{table*}[t]
  \centering
  \caption{\textbf{Controlled robustness on FLIR-Aligned and KAIST.}
  FLIR entries are mAP$_{50}$/Precision; KAIST entries are
  mAP$_{50}$/LAMR-All/Day/Night ($\downarrow$).
  Clean denotes the unmodified test set and Robust follows protocol~(ii).
  Published detector rows provide contextual comparisons; matched adaptation
  controls are reported in \cref{tab:ablation}.}
  \label{tab:main_results}
  \scriptsize \setlength{\tabcolsep}{2.2pt} \renewcommand{\arraystretch}{0.88} \setlength{\aboverulesep}{0.35ex} \setlength{\belowrulesep}{0.45ex}
  \setlength{\tabcolsep}{2.4pt}
  \begin{tabular*}{\linewidth}{
      @{\extracolsep{\fill}}lcccc@{}}
    \toprule
    \textbf{Method}
      & \textbf{FLIR Clean}
      & \textbf{FLIR Robust}
      & \textbf{KAIST Clean}
      & \textbf{KAIST Robust} \\
    \midrule
    Frozen ICAFusion~\cite{shen2024icafusion}
      & 82.8/81.2
      & 40.5/61.7
      & 76.6/12.8/12.4/5.5
      & 34.7/43.5/40.2/39.8 \\
    ProbEn~\cite{chen2022multimodal}
      & 79.6/80.5
      & 45.4/68.2
      & 74.2/12.5/11.2/5.8
      & 41.1/40.5/39.1/39.3 \\
    RSDet~\cite{11278552}
      & 83.1/83.1
      & 41.5/70.4
      & 77.1/11.7/12.1/8.3
      & 39.3/40.1/44.0/34.2 \\
    TFDet~\cite{10645696}
      & -- & --
      & 77.9/10.9/12.2/5.5
      & 36.3/42.6/45.0/35.4 \\
    C2Former~\cite{yuan2024c}
      & -- & --
      & 75.3/11.4/13.2/5.5
      & 35.1/41.6/46.3/39.4 \\
    YOLOv11-RGBT~\cite{wan2025yolov11}
      & \textbf{84.4}/82.9
      & 42.5/73.4
      & \textbf{78.3}/\textbf{9.4/10.5/4.2}
      & 37.1/39.6/37.1/40.4 \\
    UniRGB-IR~\cite{yuan2025unirgb}
      & 82.2/79.0
      & 38.5/69.4
      & 74.4/12.8/13.2/6.2
      & 35.5/42.0/40.1/38.9 \\
    \midrule
    \textbf{PG-FM LoRA}
      & 83.2/\textbf{83.4}
      & \textbf{61.3}/\textbf{75.1}
      & 77.4/12.0/12.4/5.3
      & \textbf{54.6}/\textbf{35.8/32.4/30.2} \\
    \textbf{PG-FM FiLM}
      & 82.8/82.9
      & 57.7/73.8
      & 76.6/12.7/12.4/5.4
      & 51.7/37.1/35.2/30.8 \\
    \bottomrule
  \end{tabular*}
\end{table*}

\textbf{Controlled robustness.}\quad
\Cref{tab:main_results} shows that PG-FM substantially improves
corrupted-input detection under both FiLM and LoRA while preserving the
clean performance of the corresponding frozen detector. On FLIR, LoRA and
FiLM improve robust mAP$_{50}$ by $20.8$ and $17.2$ points over baseline
ICAFusion. On KAIST, LoRA improves robust mAP$_{50}$ from $34.7$ to
$54.6$ and reduces LAMR-All from $43.5$ to $35.8$. The matched controls in
\cref{tab:ablation} separate these gains from static adaptation,
prompt-free dynamic capacity, and additional parameters.  Across three independent seeds, standard deviations for the headline
controlled and unseen-corruption evaluations remain at most $0.7$ mAP;
complete mean$\pm$std results are supplementary.

\begin{figure}[t]
  \centering
  \includegraphics[width=\linewidth]{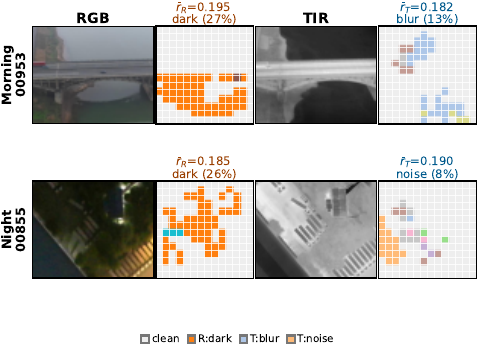}
    \caption{\textbf{Zero-shot prompt behaviour on naturally acquired
    ATR-UMOD.}
    Representative unmodified pairs remain near the clean trunk or enter
    only shallow branches, corresponding to a strong identity prior.
    Condition-stratified detection results are supplementary.}
  \label{fig:atrumod_prompt}
\end{figure}

\textbf{Zero-shot transfer to naturally acquired data.}\quad
To distinguish operator shift from acquisition shift, Cross-sim.\ replaces
our operators with the independent \texttt{imagecorruptions}
implementation~\cite{hendrycks2019robustness}, while FLIR-trained models are
evaluated without retraining or added corruption on unmodified LYNRED-MDS,
M3FD, and ATR-UMOD. PG-FM ranks first across all five domain transfer tests in
\cref{tab:external_transfer}. On ATR-UMOD, after macro-averaging category
AP$_{50}$ within each attribute, PG-FM ranks first on all six annotated
condition attributes, exceeding baseline ICAFusion by $10.8$--$18.2$ points
and Zero-Prompt by $4.7$--$7.3$ points (Supplementary). Representative
pairs in \cref{fig:atrumod_prompt} yield near-trunk or shallow-branch
prompts.

\textbf{Architecture and recovery controls.}\quad
With each detector/head frozen, PG-FM gains $17.2/16.2/17.2$ robust
mAP$_{50}$ on ICAFusion/YOLOv11-RGBT/UniRGB-IR with at most $0.3$
clean-mAP change, and exceeds matched-capacity ZP by
$8.8/8.6/5.8$ points, respectively (Supplementary). Under matched RGB
corruptions, PG-FM exceeds BioIR and PromptIR by $3.6$--$7.4$ mAP while
adding $5.46$\,GFLOPs/$2.25$\,ms, versus
$931$--$1079$\,GFLOPs/$327$--$345$\,ms for restoration
(\cref{tab:recover}). The same ordering holds on external, unseen, and mixed corruption tests
(\cref{tab:external_transfer,tab:heldout_mixed}); complete results are
supplementary.

\begin{table*}[t]
  \centering
  \captionsetup{skip=2pt}
    \caption{\textbf{Targeted adaptation, restoration, and registration
    controls} (mAP$_{50}$).
    (a) Direct reliability gating under the FLIR robust protocol.
    (b) Restore-then-detect versus PG-FM under matched RGB Gaussian noise
    (G.N.), motion blur (M.B.), and rain, with TIR clean and the detector fixed.
    (c) Isolated translation without signal corruption.
    Each panel uses a distinct evaluation protocol.}
  \label{tab:auxiliary}
  \scriptsize

  \begin{subtable}[t]{0.27\linewidth}
    \centering
    \setlength{\tabcolsep}{2.0pt}
    \renewcommand{\arraystretch}{0.82}
    \begin{tabular}{@{}lc@{}}
      \toprule
      \textbf{Adaptation} & \textbf{Rob.} \\
      \midrule
      Frozen          & 40.5 \\
      Static scalar   & 41.6 \\
      Prompt scalar   & 45.1 \\
      Prompt spatial  & 47.9 \\
      \textbf{PG-FM}  & \textbf{57.7} \\
      \bottomrule
    \end{tabular}
    \caption{Direct gating.}
    \label{tab:direct_gating}
  \end{subtable}
  \hfill
  \begin{subtable}[t]{0.30\linewidth}
    \centering
    \setlength{\tabcolsep}{2.2pt}
    \renewcommand{\arraystretch}{0.82}
    \begin{tabular}{@{}lccc@{}}
      \toprule
      \textbf{Method} & \textbf{G.N.} & \textbf{M.B.} & \textbf{Rain} \\
      \midrule
      BioIR
        & 71.0 & 71.6 & 73.6 \\
      PromptIR
        & 69.2 & 70.8 & 72.3 \\
      \textbf{PG-FM}
        & \textbf{76.5} & \textbf{75.2} & \textbf{79.7} \\
      \bottomrule
    \end{tabular}
    \caption{Restore-then-detect.}
    \label{tab:recover}
  \end{subtable}
  \hfill
  \begin{subtable}[t]{0.38\linewidth}
    \centering
    \setlength{\tabcolsep}{1.6pt}
    \renewcommand{\arraystretch}{0.82}
    \begin{tabular}{@{}lccccc@{}}
      \toprule
      \textbf{Shift} & 0 & 2 & 4 & 8 & 16 \\
      \midrule
      Frozen
        & 82.8 & 78.3 & 61.4 & 28.5 & 9.6 \\
      Zero-Prompt
        & 82.6 & 77.9 & 61.7 & 27.9 & 9.2 \\
      PG-FM
        & 82.8 & 78.3 & 62.1 & 29.9 & 10.7 \\
      \bottomrule
    \end{tabular}
    \caption{Translation shift (pixels).}
    \label{tab:misalignment}
  \end{subtable}
\end{table*}

\begin{table*}[t]
  \centering
  \caption{\textbf{Independent operator shift and zero-shot transfer to
  naturally acquired datasets} (mAP$_{50}$).
  All variants are trained only on FLIR-Aligned. Cross-sim.\ uses the
  independent \texttt{imagecorruptions} implementation on FLIR; LYNRED-MDS,
  M3FD, and ATR-UMOD are evaluated unmodified, without target-domain
  optimization or added corruption. Restoration methods follow the same
  restore-then-detect protocol, and LYNRED is scored against its IR- and
  RGB-authored annotations.}
  \label{tab:external_transfer}
  \scriptsize \setlength{\tabcolsep}{2.2pt} \renewcommand{\arraystretch}{0.88} \setlength{\aboverulesep}{0.35ex} \setlength{\belowrulesep}{0.45ex}
  \setlength{\tabcolsep}{3.0pt}
  \begin{tabular*}{\linewidth}{
      @{\extracolsep{\fill}}lccccc@{}}
    \toprule
    \textbf{Method}
      & \textbf{Cross-sim.}
      & \textbf{LYNRED-IR}
      & \textbf{LYNRED-RGB}
      & \textbf{M3FD}
      & \textbf{ATR-UMOD} \\
    \midrule

    Frozen ICAFusion
        & 42.6 & 32.2 & 29.2 & 31.4 & 19.0 \\
    BioIR$\rightarrow$ICAFusion~\cite{NEURIPS2025_73ba81c7}
        & 41.4 &  32.1 & 28.4 & 29.2 & 19.2 \\
    PromptIR$\rightarrow$ICAFusion~\cite{potlapalli2023promptir}
        & 42.2 & 31.9 & 30.0 & 30.5 & 18.4 \\
  \midrule

    Zero-Prompt
      & 48.1 & 33.3 & 32.2 & 32.3 & 27.0 \\
    Feature-RMM
      & 43.3 & 31.8 & 29.5 & 32.0 &  19.2\\
    Softmax+severity
      & 37.5 & 21.4 & 23.5 & 22.7 &  10.2\\
    Type-only
      & 50.8 & 33.4 & 31.8 & 31.7 & 25.4 \\
    Severity-only
      & 45.7 & 30.8 & 27.6 & 31.2 & 28.2 \\
    w/o $L_{\rm pair}$
      & 52.5 & 34.1 & 33.5 & 34.4 & 33.2 \\
    \textbf{PG-FM}
      & \textbf{55.2}
      & \textbf{35.9}
      & \textbf{34.4}
      & \textbf{37.0}
      & \textbf{35.7} \\
    \bottomrule
  \end{tabular*}
\end{table*}

\subsection{Prompt Geometry and Modality Adaptation}
\label{sec:robustness}

Unless noted otherwise, analyses use FiLM on FLIR-Aligned; LoRA and KAIST
results are supplementary.

\begin{figure*}[t]
  \centering
  \begin{subfigure}[b]{0.34\linewidth}
    \includegraphics[width=\linewidth]{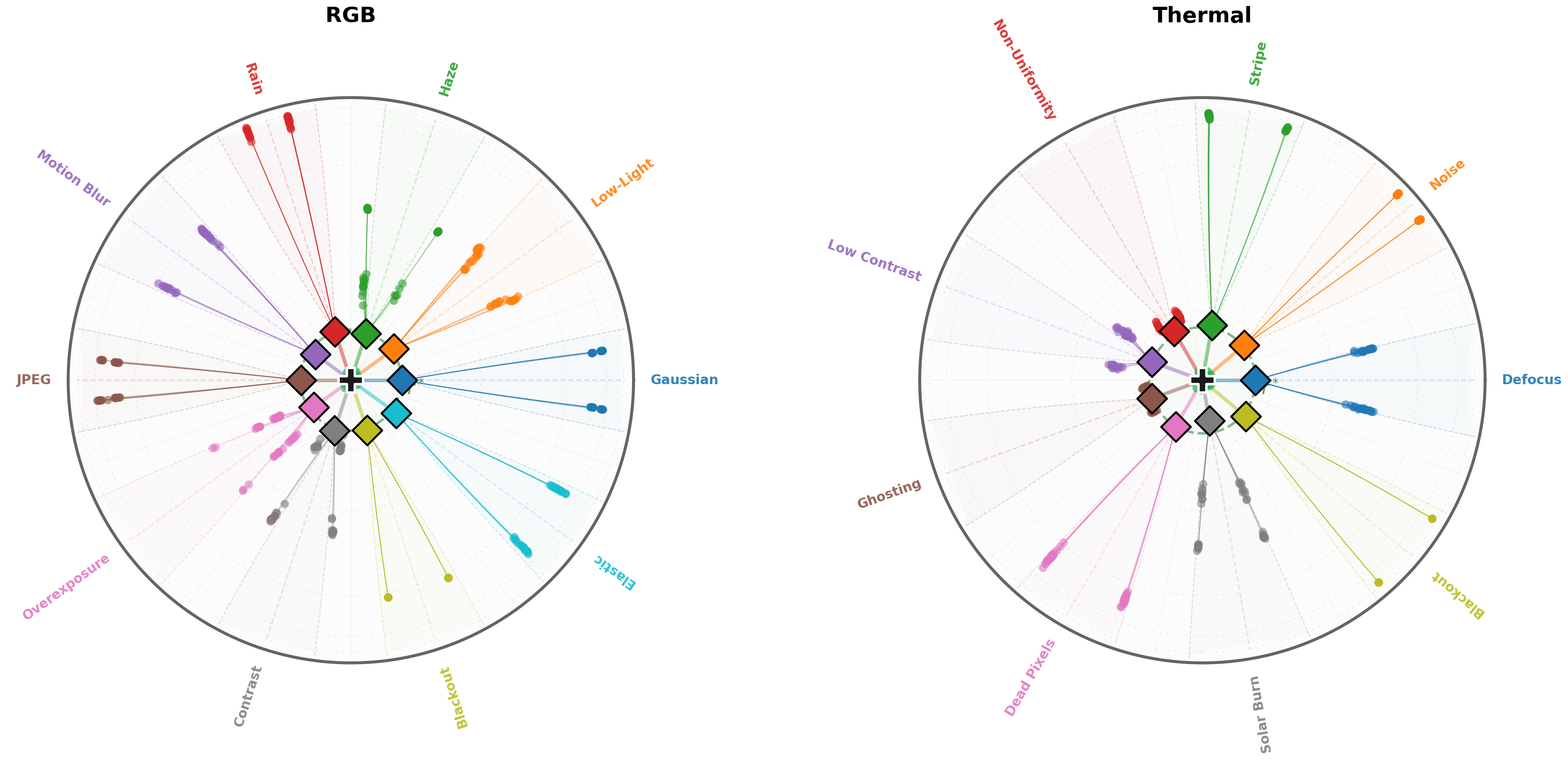}
    \caption{Clean trunk, angular corruption branches, and increasing
    geodesic depth.}
    \label{fig:stage1_ball}
  \end{subfigure}\hspace{0.02\linewidth}%
  \begin{subfigure}[b]{0.30\linewidth}
    \includegraphics[width=\linewidth]{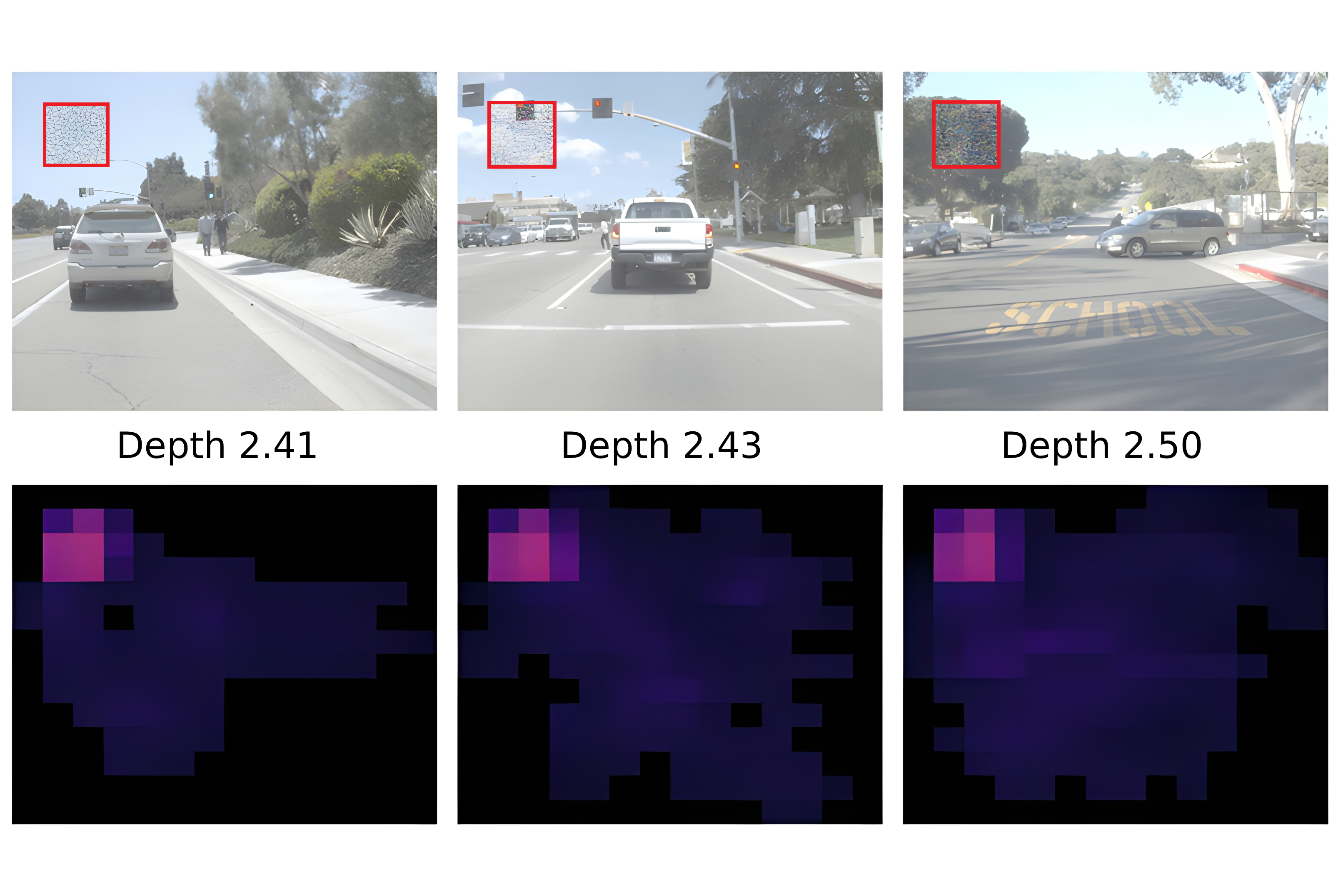}
    \caption{Corruption-identical scenes retain similar branch-relative
    depth.}
    \label{fig:stage1_montage}
  \end{subfigure}\hspace{0.02\linewidth}%
  \begin{subfigure}[b]{0.31\linewidth}
    \includegraphics[width=\linewidth]{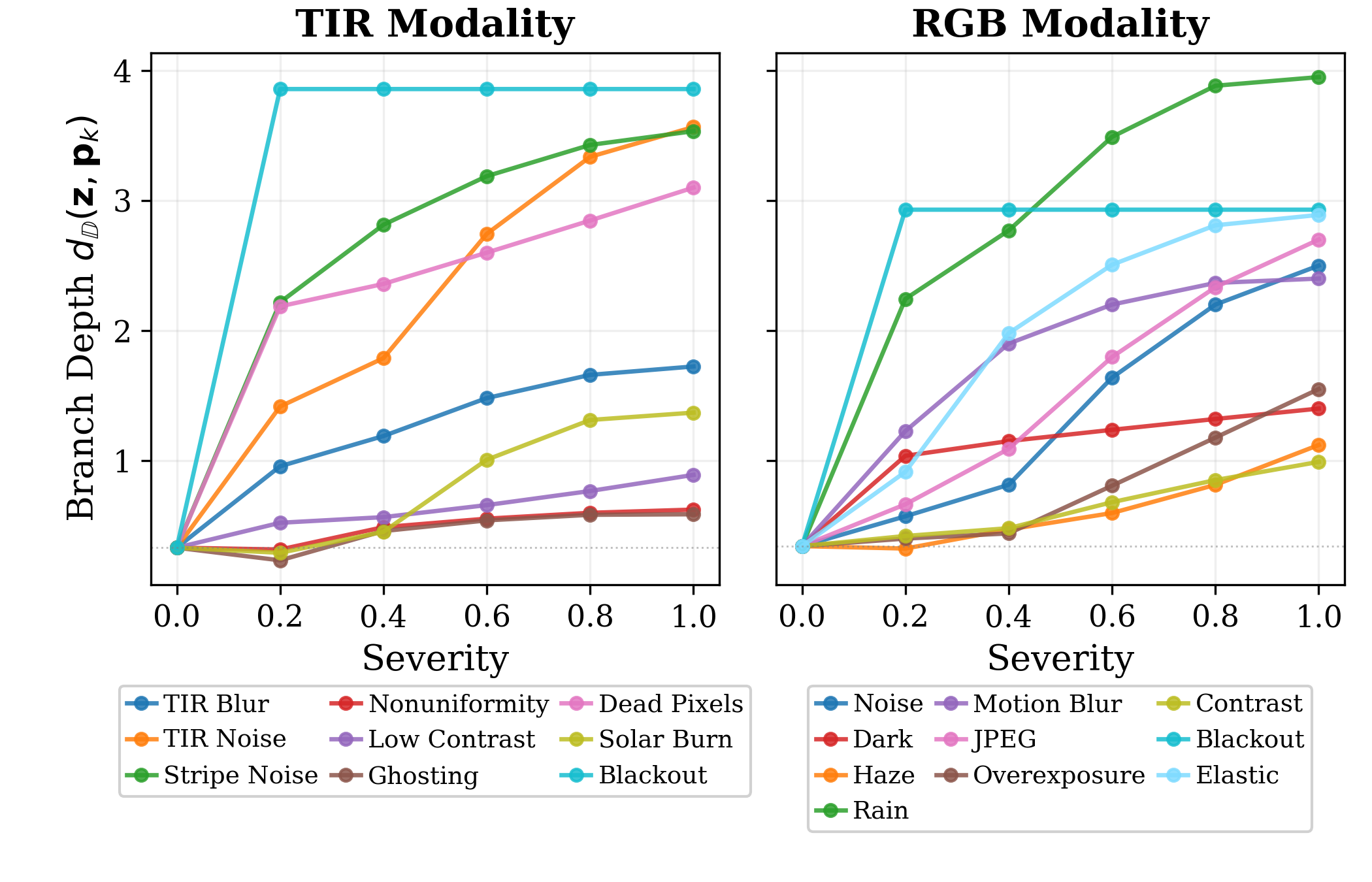}
    \caption{Branch-relative depth increases monotonically with severity.}
    \label{fig:stage1_mono}
  \end{subfigure}
  \caption{\textbf{Prompt geometry validation.}
  (a) Learned trunk-and-branch organization.
  (b) Cross-scene depth consistency.
  (c) Monotonic within-branch severity encoding in both modalities.}
  \label{fig:stage1_validation}
\end{figure*}

\textbf{Geometry and cross-type response.}\quad
\Cref{fig:stage1_validation} shows the clean trunk, angular branches,
cross-scene consistency, and monotonic within-branch depth. At fixed nominal
severity, corruption families occupy distinct pooled radii and correction
distances while the paired clean stream remains near the trunk
(Supplementary). Generated corrections cluster by type
($0.68$ silhouette) but not scene ($-0.21$), showing that pooled radius
controls identity-prior strength while the full prompt determines correction
content. Consistent with this role, removing $L_{\rm pair}$ reduces FLIR
robust mAP$_{50}$ from $57.7$ to $55.2$ and lowers all five transfer
results, linking cross-scene depth stability to downstream detection rather
than visual organization alone
(\cref{tab:ablation,tab:external_transfer}).

\textbf{Severity-resolved robustness.}\quad
Within each corruption family, increasing severity moves the
affected-modality prompts outward while the paired clean stream remains near
its clean depth. The corresponding correction distance increases with within-branch
degradation, and PG-FM consequently degrades more slowly than the Frozen,
static, and prompt-free controls. 

\begin{figure}[t]
  \centering
  \includegraphics[width=\columnwidth]
  {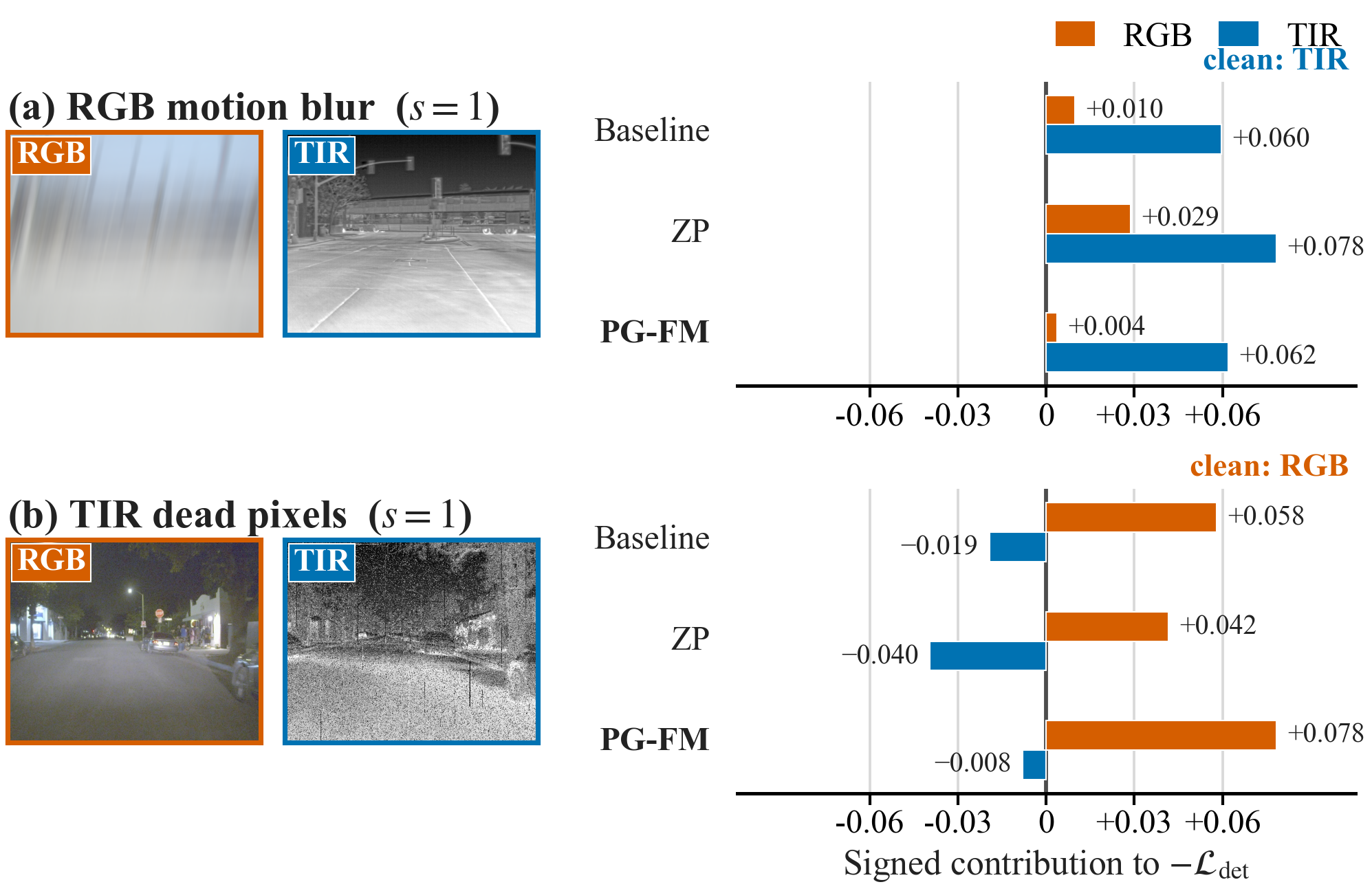}
    \caption{\textbf{Output-level modality contribution.}
    Exact signed Shapley values under severe RGB motion blur and TIR dead
    pixels. PG-FM enlarges the clean--corrupted contribution gap relative to
    Frozen and Zero-Prompt; full results are supplementary}
  \label{fig:modality_shapley}
\end{figure}

\textbf{Output-level modality contribution.}\quad
Following modality-level Shapley attribution~\cite{parcalabescu2023mm}, we
define the coalition utility as $v(S)=-\mathcal{L}_{\rm det}(S)$ and compute
\[
\phi_{\rm RGB}
=\tfrac12\!\left[v(R,0)-v(0,0)+v(R,T)-v(0,T)\right],
\]
with $\phi_{\rm TIR}$ defined symmetrically. Prompts and FiLM parameters are
inferred from the original full pair and fixed across coalitions, yielding
conditional-policy attributions under the inferred adaptation.
\Cref{fig:modality_shapley} illustrates the resulting shift toward the
uncorrupted stream in two severe cases. At $s{=}1$, the full-test-set analysis assigns $93.6/76.4\%$ of attribution magnitude to the uncorrupted TIR/RGB stream under RGB/TIR failure, versus
$90.0/75.7\%$ for Frozen and $86.1/70.3\%$ for Zero-Prompt; signed
statistics and confidence intervals are supplementary. Together, these results reveal an empirical shift in output-level reliance
toward the cleaner modality, while the identity prior regulates adaptation
capacity.

\subsection{Unseen Corruptions and Failure Boundaries}
\label{sec:generalization}

\begin{table*}[t]
  \centering
  \captionsetup{skip=2pt}
  \caption{\textbf{Unseen-corruption and mixed-corruption robustness on
  FLIR} (mAP$_{50}$).
  Unseen tests use TIR impulse noise and RGB zoom blur, barrel distortion,
  and Blackout-Blue at $s{=}1$, with the complementary modality clean.
  Mixed corruption tests use RGB Haze+Noise, RGB Dark+Rain, and TIR
  Blur+Gaussian Noise at $s{=}0.3$ per constituent, again with the
  complementary modality clean. All methods use identical pairs and the
  same frozen ICAFusion; BioIR and PromptIR restore both streams before
  detection. Full sweeps and routing analyses are supplementary.}
  \label{tab:heldout_mixed}

  \scriptsize
  \setlength{\tabcolsep}{2.2pt}
  \renewcommand{\arraystretch}{0.84}
  \setlength{\aboverulesep}{0.35ex}
  \setlength{\belowrulesep}{0.40ex}

  \begin{minipage}[t]{0.59\linewidth}
    \centering
    \setlength{\tabcolsep}{2.4pt}
    \begin{tabular}{@{}lcccc@{}}
      \toprule
      \textbf{Method}
        & \textbf{Impulse Noise}
        & \textbf{Zoom Blur}
        & \textbf{Barrel Distortion}
        & \textbf{Blackout-Blue} \\
      \midrule
      Frozen
        & 29.1 & 54.0 & 27.9 & 37.1 \\
      BioIR
        & 33.4 & 57.2 & 31.8 & 36.9 \\
      PromptIR
        & 35.1 & 56.9 & 29.2 & 36.2 \\
      Static FiLM
        & 50.0 & 61.0 & 50.1 & 52.4 \\
      Zero-Prompt
        & 50.2 & 61.2 & 49.5 & 53.7 \\
      \textbf{PG-FM}
        & \textbf{54.3}
        & \textbf{73.5}
        & \textbf{69.4}
        & \textbf{63.2} \\
      \bottomrule
    \end{tabular}\\[-1pt]
    \textbf{(a) Unseen corruption types}
  \end{minipage}
  \hfill
  \begin{minipage}[t]{0.39\linewidth}
    \centering
    \setlength{\tabcolsep}{2.4pt}
    \begin{tabular}{@{}lccc@{}}
      \toprule
      \textbf{Method}
        & \textbf{H+N}
        & \textbf{D+R}
        & \textbf{TIR-B+GN} \\
      \midrule
      Frozen
        & 46.9 & 73.4 & 35.3 \\
      BioIR
        & 50.2 & 73.1 & 36.6 \\
      PromptIR
        & 52.9 & 74.0 & 37.2 \\
      Zero-Prompt
        & 68.0 & 73.0 & 44.2 \\        
      \textbf{PG-FM}
        & \textbf{74.2}
        & \textbf{76.5}
        & \textbf{55.9} \\
      \bottomrule
    \end{tabular}\\[-1pt]
    \textbf{(b) Mixed corruption tests}
  \end{minipage}
\end{table*}

\textbf{Unseen corruption and mixed corruption tests.}\quad
PG-FM remains strongest across all tested unseen operators and mixed
corruption tests in \cref{tab:heldout_mixed}. Supplementary routing sweeps
show that structurally related unseen corruptions map to nearby learned
branches with depth increasing with severity. Under mixed corruption, the
prompt is inferred jointly from spatial--energy evidence and follows the
dominant signature at a depth comparable to the stronger constituent.
Together, these results demonstrate transfer to unseen operators and
robustness under the tested corruption mixtures.

\textbf{Registration sensitivity.}\quad
Under isolated translation, prompt embeddings remain near the clean trunk
and the RMM stays close to identity; consequently, PG-FM tracks the Frozen
detector rather than compensating for geometric displacement
(\cref{tab:misalignment}). Performance degrades with severe correspondence
loss for all methods, making explicit registration complementary to the
signal-adaptation mechanism.

\subsection{Controlled Ablations}
\label{sec:ablations}

Unless noted otherwise, ablations use FiLM on FLIR-Aligned; LoRA, KAIST,
and per-corruption results are supplementary.

\textbf{Ablation protocol.}\quad
All variants share the frozen detector, corruption sampling, fusion targets,
training schedule, and matched seeds. For conditioning comparisons,
trainable variants use the same fixed identity-centred $\ell_2$ penalty and
are scaled to the full two-PG-plus-RMM parameter budget. \emph{Zero-Prompt}
widens the RMM to this budget but supplies a constant input.
\emph{Feature-RMM} bypasses Stage~1 by pooling and concatenating frozen
multi-scale RGB/TIR features and feeding their projection directly to the
same RMM. \emph{Type-only} retains the clean-boundary and
type-classification losses but removes severity ranking and cross-scene
consistency, whereas \emph{Severity-only} removes type classification while
retaining the boundary, ranking, and consistency losses. \emph{Scalar} and
\emph{Softmax+severity} use the same diagnostic inputs and supervision but
replace the geometric prompt with a scalar regressor or a factorized
classifier--regressor. Finally, PG-$\ell_2$ retains the complete prompt with
constant identity regularisation, isolating the reliability-weighted
identity regularisation used by full PG-FM.

\begin{table*}[t]
  \centering
  \caption{\textbf{Matched conditioning and component ablations on
  FLIR-Aligned and KAIST.}
  FLIR entries are mAP$_{50}$/Precision; KAIST entries are
  mAP$_{50}$/LAMR-All/Day/Night ($\downarrow$).
  L/F denote LoRA/FiLM. All variants use the same frozen detector and
  corruption protocol; Zero-Prompt uses the unified capacity-matched
  implementation. Additional prompt, backend, and per-corruption controls
  are supplementary.}
  \label{tab:ablation}
  \scriptsize \setlength{\tabcolsep}{2.2pt} \renewcommand{\arraystretch}{0.88} \setlength{\aboverulesep}{0.35ex} \setlength{\belowrulesep}{0.45ex}
  \setlength{\tabcolsep}{1.8pt}
  \begin{tabular*}{\linewidth}{
      @{\extracolsep{\fill}}clcccc@{}}
    \toprule
    \textbf{B}
      & \textbf{Variant}
      & \textbf{FLIR Clean}
      & \textbf{FLIR Robust}
      & \textbf{KAIST Clean}
      & \textbf{KAIST Robust} \\
    \midrule

    L & Static
      & 82.6/80.9
      & 57.7/70.8
      & 76.2/13.1/11.9/5.3
      & 47.8/41.7/39.9/34.4 \\
    L & Zero-Prompt
      & 82.4/81.4
      & 59.8/72.2
      & 75.4/13.3/11.9/6.7
      & 51.7/39.4/38.5/31.3 \\
    L & \textbf{PG-FM}
      & 83.2/83.4
      & \textbf{61.3}/75.1
      & 77.4/12.0/12.4/5.3
      & \textbf{54.6}/35.8/32.4/30.2 \\

    \midrule

    F & Static
      & 82.6/81.1
      & 48.8/68.8
      & 76.1/13.3/13.5/5.9
      & 46.4/39.1/38.5/36.4 \\
    F & w/o Energy Token
      & 81.0/78.6
      & 49.3/65.5
      & 74.5/15.9/17.5/10.4
      & 45.4/39.5/39.1/37.9 \\
    F & w/o $L_{\rm pair}$
      & 82.9/82.5
      & 55.2/71.4
      & 76.5/12.8/13.1/5.5
      & 49.7/38.3/37.0/34.5 \\
    F & Zero-Prompt
      & 82.6/80.9
      & 48.9/68.3
      & 76.3/13.3/13.8/6.1
      & 46.2/39.5/39.2/36.8 \\
    F & Euclidean-polar
      & 80.8/82.0
      & 52.9/66.3
      & 75.0/14.3/13.3/7.2
      & 42.3/38.5/41.5/38.3 \\
    F & PG-$\ell_2$
      & 82.8/83.0
      & 54.8/71.2
      & 76.5/12.9/12.7/5.5
      & 49.1/38.6/37.1/34.9 \\
    F & \textbf{PG-FM}
      & 82.8/82.9
      & \textbf{57.7}/73.8
      & 76.6/12.7/12.4/5.4
      & \textbf{51.7}/37.1/35.2/30.8 \\
    \bottomrule
  \end{tabular*}
\end{table*}

\textbf{Conditioning and component controls.}\quad
\Cref{tab:ablation,tab:external_transfer} assign one matched control to each
claim. Static FiLM/LoRA test input-independent PEFT; Zero-Prompt isolates
prompt-free dynamic capacity; Feature-RMM tests direct conditioning on
multi-scale features; Type-only, Severity-only, and Softmax+severity isolate
the joint type--severity factorization; Euclidean-polar changes only the
metric; w/o Energy Token removes absolute-signal cues; w/o
$L_{\rm pair}$ removes cross-scene depth consistency; and PG-$\ell_2$
replaces reliability-weighted identity regularisation with a constant
penalty. Full PG-FM is strongest across in-domain and transfer settings.
Type-only remains near Zero-Prompt, whereas Severity-only transfers less
consistently. Removing the Energy Token weakens absolute-signal calibration
and clean performance, while removing $L_{\rm pair}$ weakens cross-scene
depth stability and external transfer. PG-$\ell_2$ remains below the full model. At matched $8$D,
Euclidean-polar reaches $52.9$ robust mAP versus $57.7$ for Poincar\'e;
$16$D/$32$D Euclidean reaches $53.7/57.2$, supporting a compact
fixed-dimensional advantage (Supplementary).

\textbf{Direct reliability gating.}\quad
Static scalar gates learn one input-independent RGB/TIR weight at each
fusion level, while prompt scalar and spatial gates predict per-level
weights or $14{\times}14$ reliability maps from the Stage~1 prompts.
Although increasingly expressive gating improves robustness, all variants
remain below PG-FM (\cref{tab:direct_gating}), showing that reliability
diagnosis alone does not replace layer- and channel-specific correction.

\section{Conclusion}
PG-FM provides signal-quality-conditioned fusion adaptation, improving robustness while preserving clean accuracy with modest overhead. Across controlled corruptions and naturally acquired external datasets, it consistently improves RGB--T detection and generalizes without retraining; fault-labeled natural sensor-failure validation remains future work.
{
    \small
    \bibliographystyle{ieeenat_fullname}
    \bibliography{main}

@String(IJCV  = {Int. J. Comput. Vis.})

@String(CVPR  = {IEEE Conf. Comput. Vis. Pattern Recog.})

@String(ICCV  = {Int. Conf. Comput. Vis.})

@String(ECCV  = {Eur. Conf. Comput. Vis.})

@String(NeurIPS = {Adv. Neural Inform. Process. Syst.})

@String(ICLR  = {Int. Conf. Learn. Represent.})

@String(AAAI  = {AAAI})

@String(ICIP  = {IEEE Int. Conf. Image Process.})

@String(TOG   = {ACM Trans. Graph.})

@String(ACMMM = {ACM Int. Conf. Multimedia})

@String(PR    = {Pattern Recognition})

@String(IJCV  = {IJCV})

@String(CVPR  = {CVPR})

@String(ICCV  = {ICCV})

@String(ECCV  = {ECCV})

@String(NeurIPS = {NeurIPS})

@String(ICLR  = {ICLR})

@String(ICIP  = {ICIP})

@String(TOG   = {ACM TOG})

@String(ACMMM = {ACM MM})

@String(PR    = {PR})

@inproceedings{NEURIPS2024_f29369d1,
  author    = {Aithal, Sumukh K and Maini, Pratyush and Lipton, Zachary C. and Kolter, J. Zico},
  booktitle = NeurIPS,
  title     = {Understanding Hallucinations in Diffusion Models through Mode Interpolation},
  pages     = {134614--134644},
  volume    = {37},
  year      = {2024},
  doi       = {10.52202/079017-4278}
}

@INPROCEEDINGS{10030129,
  author={Babu, Nithin C and Kannan, Vignesh and Soundararajan, Rajiv},
  booktitle={2023 IEEE/CVF Winter Conference on Applications of Computer Vision (WACV)}, 
  title={No Reference Opinion Unaware Quality Assessment of Authentically Distorted Images}, 
  year={2023},
  volume={},
  number={},
  pages={2458-2467},
  doi={10.1109/WACV56688.2023.00249}}

@inproceedings{chen2022multimodal,
  title     = {Multimodal Object Detection via Probabilistic Ensembling},
  author    = {Chen, Yi-Ting and Shi, Jinghao and Ye, Zelin and Mertz, Christoph and Ramanan, Deva and Kong, Shu},
  booktitle = ECCV,
  pages     = {139--158},
  year      = {2022},
  doi       = {10.1007/978-3-031-20077-9_9}
}

@inproceedings{khrulkov2020hyperbolic,
  title={Hyperbolic image embeddings},
  author={Khrulkov, Valentin and Mirvakhabova, Leyla and Ustinova, Evgeniya and Oseledets, Ivan and Lempitsky, Victor},
  booktitle={2020 IEEE/CVF conference on computer vision and pattern recognition (CVPR)},
  pages={6417--6427},
  year={2020},
  organization={IEEE}
}

@inproceedings{zhang2020multispectral,
  title={Multispectral fusion for object detection with cyclic fuse-and-refine blocks},
  author={Zhang, Heng and Fromont, Elisa and Lefevre, S{\'e}bastien and Avignon, Bruno},
  booktitle={2020 IEEE International conference on image processing (ICIP)},
  pages={276--280},
  year={2020},
  organization={IEEE}
}

@inproceedings{ganea2018hyperbolic,
  author    = {Ganea, Octavian and Becigneul, Gary and Hofmann, Thomas},
  booktitle = NeurIPS,
  title     = {Hyperbolic Neural Networks},
  volume    = {31},
  year      = {2018}
}

@article{gao2024clip,
  title   = {{CLIP-Adapter}: Better Vision-Language Models with Feature Adapters},
  author  = {Gao, Peng and Geng, Shijie and Zhang, Renrui and Ma, Teli and Fang, Rongyao and Zhang, Yongfeng and Li, Hongsheng and Qiao, Yu},
  journal = IJCV,
  volume  = {132},
  number  = {2},
  pages   = {581--595},
  year    = {2024},
  doi     = {10.1007/s11263-023-01891-x}
}

@article{garg2006photorealistic,
  title   = {Photorealistic Rendering of Rain Streaks},
  author  = {Garg, Kshitiz and Nayar, Shree K.},
  journal = TOG,
  volume  = {25},
  number  = {3},
  pages   = {996--1002},
  year    = {2006},
  doi     = {10.1145/1141911.1141985}
}

@inproceedings{he2023multispectral,
  title     = {Multispectral Object Detection via Cross-Modal Conflict-Aware Learning},
  author    = {He, Xiao and Tang, Chang and Zou, Xin and Zhang, Wei},
  booktitle = ACMMM,
  pages     = {1465--1474},
  year      = {2023},
  doi       = {10.1145/3581783.3612651}
}

@inproceedings{hendrycks2019robustness,
  title     = {Benchmarking Neural Network Robustness to Common Corruptions and Perturbations},
  author    = {Hendrycks, Dan and Dietterich, Thomas},
  booktitle = ICLR,
  year      = {2019}
}

@inproceedings{hu2022lora,
  title     = {{LoRA}: Low-Rank Adaptation of Large Language Models},
  author    = {Hu, Edward J. and Shen, Yelong and Wallis, Phillip and Allen-Zhu, Zeyuan and Li, Yuanzhi and Wang, Shean and Wang, Lu and Chen, Weizhu},
  booktitle = ICLR,
  year      = {2022}
}

@INPROCEEDINGS{7298706,
  author    = {Hwang, Soonmin and Park, Jaesik and Kim, Namil and Choi, Yukyung and Kweon, In So},
  booktitle = CVPR,
  title     = {Multispectral Pedestrian Detection: Benchmark Dataset and Baseline},
  year      = {2015},
  pages     = {1037--1045},
  doi       = {10.1109/CVPR.2015.7298706}
}

@inproceedings{kim2025exploiting,
  title={Exploiting diffusion prior for task-driven image restoration},
  author={Kim, Jaeha and Oh, Junghun and Lee, Kyoung Mu},
  booktitle={2025 IEEE/CVF International Conference on Computer Vision (ICCV)},
  pages={10151--10161},
  year={2025},
  organization={IEEE}
}

@article{kim2025hallugen,
  title={HalluGen: Synthesizing Realistic and Controllable Hallucinations for Evaluating Image Restoration},
  author={Kim, Seunghoi and Tregidgo, Henry FJ and Jin, Chen and Figini, Matteo and Alexander, Daniel C},
  journal={arXiv preprint arXiv:2512.03345},
  year={2025}
}

@Article{s23208424,
AUTHOR = {Li, Mingqing and Wang, Yuqing and Sun, Haijiang},
TITLE = {Single-Frame Infrared Image Non-Uniformity Correction Based on Wavelet Domain Noise Separation},
JOURNAL = {Sensors},
VOLUME = {23},
YEAR = {2023},
NUMBER = {20},
ARTICLE-NUMBER = {8424},
URL = {https://www.mdpi.com/1424-8220/23/20/8424},
PubMedID = {37896518},
ISSN = {1424-8220},
DOI = {10.3390/s23208424}
}

@article{li2024flat,
  title   = {{Flat-LoRA}: Low-Rank Adaptation over a Flat Loss Landscape},
  author  = {Li, Tao and He, Zhengbao and Li, Yujun and Wang, Yasheng and Shang, Lifeng and Huang, Xiaolin},
  journal = {arXiv preprint arXiv:2409.14396},
  year    = {2024}
}

@inproceedings{liu2025deal,
  title     = {{DEAL}: Data-Efficient Adversarial Learning for High-Quality Infrared Imaging},
  author    = {Liu, Zhu and Wang, Zijun and Liu, Jinyuan and Meng, Fanqi and Ma, Long and Liu, Risheng},
  booktitle = CVPR,
  pages     = {28198--28207},
  year      = {2025}
}

@inproceedings{ma2022multimodal,
  title     = {Are Multimodal Transformers Robust to Missing Modality?},
  author    = {Ma, Mengmeng and Ren, Jian and Zhao, Long and Testuggine, Davide and Peng, Xi},
  booktitle = CVPR,
  pages     = {18177--18186},
  year      = {2022},
  doi       = {10.1109/CVPR52688.2022.01764}
}

@inproceedings{mai2025lessons,
  title={Lessons and insights from a unifying study of parameter-efficient fine-tuning (peft) in visual recognition},
  author={Mai, Zheda and Zhang, Ping and Tu, Cheng-Hao and Chen, Hong-You and Nguyen, Quang-Huy and Zhang, Li and Chao, Wei-Lun},
  booktitle={2025 IEEE/CVF Conference on Computer Vision and Pattern Recognition (CVPR)},
  pages={14845--14857},
  year={2025},
  organization={IEEE}
}

@article{nickel2017poincare,
  title={Poincar{\'e} embeddings for learning hierarchical representations},
  author={Nickel, Maximillian and Kiela, Douwe},
  journal={Advances in neural information processing systems},
  volume={30},
  year={2017}
}

@inproceedings{perez2018film,
  title={Film: Visual reasoning with a general conditioning layer},
  author={Perez, Ethan and Strub, Florian and De Vries, Harm and Dumoulin, Vincent and Courville, Aaron},
  booktitle={Proceedings of the AAAI conference on artificial intelligence},
  volume={32},
  number={1},
  year={2018}
}

@inproceedings{pfeiffer2020adapterhub,
  title={Adapterhub: A framework for adapting transformers},
  author={Pfeiffer, Jonas and R{\"u}ckl{\'e}, Andreas and Poth, Clifton and Kamath, Aishwarya and Vuli{\'c}, Ivan and Ruder, Sebastian and Cho, Kyunghyun and Gurevych, Iryna},
  booktitle={Proceedings of the 2020 conference on empirical methods in natural language processing: system demonstrations},
  pages={46--54},
  year={2020}
}

@article{shen2024icafusion,
  title   = {{ICAFusion}: Iterative Cross-Attention Guided Feature Fusion for Multispectral Object Detection},
  author  = {Shen, Jifeng and Chen, Yifei and Liu, Yue and Zuo, Xin and Fan, Heng and Yang, Wankou},
  journal = PR,
  volume  = {145},
  pages   = {109913},
  year    = {2024},
  doi     = {10.1016/j.patcog.2023.109913}
}

@inproceedings{shi2024transformer,
  title     = {Transformer-Based No-Reference Image Quality Assessment via Supervised Contrastive Learning},
  author    = {Shi, Jinsong and Gao, Pan and Qin, Jie},
  booktitle = AAAI,
  volume    = {38},
  pages     = {4829--4837},
  year      = {2024},
  doi       = {10.1609/aaai.v38i5.28285}
}

@INPROCEEDINGS{9156687,
  author    = {Su, Shaolin and Yan, Qingsen and Zhu, Yu and Zhang, Cheng and Ge, Xin and Sun, Jinqiu and Zhang, Yanning},
  booktitle = CVPR,
  title     = {Blindly Assess Image Quality in the Wild Guided by a Self-Adaptive Hyper Network},
  year      = {2020},
  pages     = {3664--3673},
  doi       = {10.1109/CVPR42600.2020.00372}
}

@inproceedings{ermolov2022hyperbolic,
  title={Hyperbolic vision transformers: Combining improvements in metric learning},
  author={Ermolov, Aleksandr and Mirvakhabova, Leyla and Khrulkov, Valentin and Sebe, Nicu and Oseledets, Ivan},
  booktitle={2022 IEEE/CVF Conference on Computer Vision and Pattern Recognition (CVPR)},
  pages={7399--7409},
  year={2022},
  organization={IEEE}
}

@inproceedings{NEURIPS2022_1cac8326,
 author = {Sun, Shangquan and Ren, Wenqi and Wang, Tao and Cao, Xiaochun},
 booktitle = {Advances in Neural Information Processing Systems},
 doi = {10.52202/068431-0322},
 editor = {S. Koyejo and S. Mohamed and A. Agarwal and D. Belgrave and K. Cho and A. Oh},
 pages = {4461--4474},
 publisher = {Curran Associates, Inc.},
 title = {Rethinking Image Restoration for Object Detection},
 url = {https://proceedings.neurips.cc/paper_files/paper/2022/file/1cac8326ce3fbe79171db9754211530c-Paper-Conference.pdf},
 volume = {35},
 year = {2022}
}

@inproceedings{taori2020measuring,
  title     = {Measuring Robustness to Natural Distribution Shifts in Image Classification},
  author    = {Taori, Rohan and Dave, Achal and Shankar, Vaishaal and Carlini, Nicholas and Recht, Benjamin and Schmidt, Ludwig},
  booktitle = NeurIPS,
  volume    = {33},
  pages     = {18583--18599},
  year      = {2020}
}

@article{TIAN20251,
  title   = {Learning a Robust {RGB}-Thermal Detector for Extreme Modality Imbalance},
  author  = {Tian, Chao and Yang, Chao and Zhu, Guoqing and Wang, Qiang and He, Zhenyu},
  journal = {Pattern Recognition Letters},
  volume  = {196},
  pages   = {1--8},
  year    = {2025},
  doi     = {10.1016/j.patrec.2025.05.005}
}

@article{wan2025yolov11,
  title   = {{YOLOv11-RGBT}: Towards a Comprehensive Single-Stage Multispectral Object Detection Framework},
  author  = {Wan, Dahang and Lu, Rongsheng and Fang, Yang and Lang, Xianli and Shu, Shuangbao and Chen, Jingjing and Shen, Siyuan and Xu, Ting and Ye, Zecong},
  journal = {arXiv preprint arXiv:2506.14696},
  year    = {2025}
}

@ARTICLE{10817087,
  author={Wang, Qingwang and Sun, Yuxuan and Chi, Yongke and Shen, Tao},
  journal={IEEE Journal of Selected Topics in Applied Earth Observations and Remote Sensing}, 
  title={RGB-T Object Detection With Failure Scenarios}, 
  year={2025},
  volume={18},
  number={},
  pages={3000-3010},
  doi={10.1109/JSTARS.2024.3523408}}

@INPROCEEDINGS{10655501,
  author    = {Xu, Kangmin and Liao, Liang and Xiao, Jing and Chen, Chaofeng and Wu, Haoning and Yan, Qiong and Lin, Weisi},
  booktitle = CVPR,
  title     = {Boosting Image Quality Assessment Through Efficient Transformer Adaptation with Local Feature Enhancement},
  year      = {2024},
  pages     = {2662--2672},
  doi       = {10.1109/CVPR52733.2024.00257}
}

@inproceedings{yuan2025unirgb,
  title     = {{UniRGB-IR}: A Unified Framework for Visible-Infrared Semantic Tasks via Adapter Tuning},
  author    = {Yuan, Maoxun and Cui, Bo and Zhao, Tianyi and Wang, Jiayi and Fu, Shan and Yang, Xue and Wei, Xingxing},
  booktitle = ACMMM,
  pages     = {2409--2418},
  year      = {2025}
}

@article{yuan2024c,
  title   = {{C$^2$Former}: Calibrated and Complementary Transformer for {RGB}-Infrared Object Detection},
  author  = {Yuan, Maoxun and Wei, Xingxing},
  journal = {IEEE Transactions on Geoscience and Remote Sensing},
  volume  = {62},
  pages   = {1--12},
  year    = {2024},
  doi     = {10.1109/TGRS.2024.3376819}
}

@ARTICLE{10645696,
  author={Zhang, Xue and Zhang, Xiaohan and Wang, Jiangtao and Ying, Jiacheng and Sheng, Zehua and Yu, Heng and Li, Chunguang and Shen, Hui-Liang},
  journal={IEEE Transactions on Neural Networks and Learning Systems}, 
  title={TFDet: Target-Aware Fusion for RGB-T Pedestrian Detection}, 
  year={2025},
  volume={36},
  number={7},
  pages={13276-13290},
  doi={10.1109/TNNLS.2024.3443455}}

@ARTICLE{11133673,
  author={Zhang, Yusi and Xie, Weiying and Hui, Tianlin and Li, Daixun and Zhang, Jiaqing and Lei, Jie and Li, Yunsong and Fang, Leyuan},
  journal={IEEE Transactions on Circuits and Systems for Video Technology}, 
  title={LoME: LoRA-Driven Multimodal Extractor for RGB-X Vision Tasks}, 
  year={2026},
  volume={36},
  number={2},
  pages={1487-1500},
  doi={10.1109/TCSVT.2025.3601340}}

@ARTICLE{11278552,
  author={Zhao, Tianyi and Yuan, Maoxun and Jiang, Feng and Wang, Nan and Wei, Xingxing},
  journal={IEEE Transactions on Intelligent Transportation Systems}, 
  title={Removal Then Selection: A Coarse-to-Fine Fusion Perspective for RGB-Infrared Object Detection}, 
  year={2026},
  volume={27},
  number={2},
  pages={2504-2519},
  doi={10.1109/TITS.2025.3638627}}

@InProceedings{Zhou_2024_CVPR,
    author    = {Zhou, Xin and Liang, Dingkang and Xu, Wei and Zhu, Xingkui and Xu, Yihan and Zou, Zhikang and Bai, Xiang},
    title     = {Dynamic Adapter Meets Prompt Tuning: Parameter-Efficient Transfer Learning for Point Cloud Analysis},
    booktitle = {Proceedings of the IEEE/CVF Conference on Computer Vision and Pattern Recognition (CVPR)},
    month     = {June},
    year      = {2024},
    pages     = {14707-14717}
}

@inproceedings{NEURIPS2025_73ba81c7,
 author = {Cui, Yuning and Ren, Wenqi and Knoll, Alois},
 booktitle = {Advances in Neural Information Processing Systems},
 doi = {10.52202/085713-2689},
 editor = {D. Belgrave and C. Zhang and H. Lin and R. Pascanu and P. Koniusz and M. Ghassemi and N. Chen},
 pages = {80452--80481},
 publisher = {Curran Associates, Inc.},
 title = {Bio-Inspired Image Restoration},
 url = {https://proceedings.neurips.cc/paper_files/paper/2025/file/73ba81c7b25134a559c8a9c39ec1a4c3-Paper-Conference.pdf},
 volume = {38, Main Conference},
 year = {2025}
}

@inproceedings{potlapalli2023promptir,
  title     = {{PromptIR}: Prompting for All-in-One Image Restoration},
  author    = {Potlapalli, Vaishnav and Zamir, Syed Waqas and Khan, Salman H. and Khan, Fahad Shahbaz},
  booktitle = NeurIPS,
  volume    = {36},
  pages     = {71275--71293},
  year      = {2023}
}

@ARTICLE{11533396,
  author={Arbez, Loïc and Matias, Jessy and Brenière, Xavier and Chanussot, Jocelyn and Phlypo, Ronald},
  journal={IEEE Data Descriptions}, 
  title={Descriptor: LYNRED Mobility Dataset Multimodal Detection Subset (LYNRED-MDS)}, 
  year={2026},
  volume={3},
  number={},
  pages={413-421},
  doi={10.1109/IEEEDATA.2026.3695478}}

@inproceedings{liu2022target,
  title={Target-aware dual adversarial learning and a multi-scenario multi-modality benchmark to fuse infrared and visible for object detection},
  author={Liu, Jinyuan and Fan, Xin and Huang, Zhanbo and Wu, Guanyao and Liu, Risheng and Zhong, Wei and Luo, Zhongxuan},
  booktitle={2022 IEEE/CVF conference on computer vision and pattern recognition (CVPR)},
  pages={5792--5801},
  year={2022},
  organization={IEEE}
}

@inproceedings{chen2025fusion,
  title={Fusion meets diverse conditions: A high-diversity benchmark and baseline for uav-based multimodal object detection with condition cues},
  author={Chen, Chen and Bin, Kangcheng and Hu, Ting and Qi, Jiahao and Liu, Xingyue and Liu, Tianpeng and Liu, Zhen and Liu, Yongxiang and Zhong, Ping},
  booktitle={2025 IEEE/CVF International Conference on Computer Vision (ICCV)},
  pages={27958--27967},
  year={2025},
  organization={IEEE}
}

@inproceedings{parcalabescu2023mm,
  title={Mm-shap: A performance-agnostic metric for measuring multimodal contributions in vision and language models \& tasks},
  author={Parcalabescu, Letitia and Frank, Anette},
  booktitle={Proceedings of the 61st Annual Meeting of the Association for Computational Linguistics (Volume 1: Long Papers)},
  pages={4032--4059},
  year={2023}
}

@book{amon2009evaluation,
  title={Evaluation of image quality of thermal imagers used by the fire service},
  author={Amon, Francine and Lock, Andrew},
  year={2009},
  publisher={US Department of Commerce, National Institute of Standards and Technology}
}

@article{li2015intrusion,
  title={Intrusion detection and cooperative tracking using PTZ network thermal imagers},
  author={Li, Zhenghao and Yang, Junying and Han, Peng and Ran, Yang and Chai, Zhi},
  journal={International Journal of Distributed Sensor Networks},
  volume={11},
  number={7},
  pages={130103},
  year={2015},
  publisher={SAGE Publications, Sage UK: London, England}
}
}

\end{document}